\documentclass[
]{ceurart}
\usepackage{xspace}
\usepackage{acronym}
\usepackage{listings}
\usepackage{multirow}
\def\ie{\emph{i.e.}} 

\begin{document}
\copyrightyear{2024}
\copyrightclause{Copyright for this paper by its authors.
  Use permitted under Creative Commons License Attribution 4.0
  International (CC BY 4.0).}

\conference{De-Factify 4: 4rd Workshop on Multimodal Fact Checking and Hate Speech Detection, co-located with AAAI 2025. 2025 Philadelphia, Pennsylvania, USA}

\title{Scalable Framework for Classifying AI-Generated Content Across Modalities}

\author[1]{Anh-Kiet Duong}[%
orcid=0009-0005-0230-6104,
email=anh.duong@univ-lr.fr,
url=https://ffyyytt.github.io/,
]
\cormark[1]
\address[1]{L3i Laboratory, La Rochelle University,
  Avenue Michel Crépeau, 17042 La Rochelle Cedex 1 - France}

\author[1]{Petra Gomez-Krämer}[%
orcid=0000-0002-5515-7828,
email=petra.gomez@univ-lr.fr,
]

\cortext[1]{Corresponding author.}

\begin{abstract}
  The rapid growth of generative AI technologies has heightened the importance of effectively distinguishing between human and AI-generated content, as well as classifying outputs from diverse generative models. This paper presents a scalable framework that integrates perceptual hashing, similarity measurement, and pseudo-labeling to address these challenges. Our method enables the incorporation of new generative models without retraining, ensuring adaptability and robustness in dynamic scenarios. Comprehensive evaluations on the Defactify4 dataset demonstrate competitive performance in text and image classification tasks, achieving high accuracy across both distinguishing human and AI-generated content and classifying among generative methods. These results highlight the framework’s potential for real-world applications as generative AI continues to evolve. Source codes are publicly available at \url{https://github.com/ffyyytt/defactify4}.
\end{abstract}


\maketitle

\section{Introduction}
The proliferation of generative artificial intelligence (AI) technologies has introduced new societal challenges, particularly in distinguishing between human-generated and AI-generated content \cite{wang2023survey}. The increasing sophistication of these models allows them to produce text and images that are often indistinguishable from human creations, raising concerns over their misuse in spreading misinformation, generating fake news, and creating deceptive media. Beyond this, the diversity of generative AI models, each employing distinct architectures and techniques highlights the importance of not only identifying AI-generated content but also classifying it by its source. Such classifications enable deeper forensic insights and are critical for trust-building in applications like content moderation, digital forensics, and fact-checking.

A critical aspect of this challenge is the need to differentiate between the various generative methods themselves. This distinction is essential for assessing the relative difficulty of detecting different approaches and understanding which models are more or less susceptible to detection. Without this understanding, it would be difficult to evaluate the effectiveness of detection systems and their ability to keep pace with emerging AI technologies \cite{ojha2023towards}.

Compounding these challenges is the relentless growth in the number and variety of generative models. The landscape of generative AI is far from static, with new methods continually emerging and pushing the boundaries of realism and creativity. This rapid evolution necessitates the development of adaptable classification frameworks capable of integrating new generative models without requiring costly and exhaustive retraining \cite{lin2024detecting}. Addressing this issue is crucial for building sustainable and scalable detection systems.

To address these challenges, this paper proposes a novel approach tailored for classifying and detecting AI-generated content in both text and image domains. The key contributions of this work are:  
\begin{itemize}  
    \item A method that adapts to new generative models by incorporating their features into the classification pipeline without retraining
    \item The evaluation of contrastive learning for incremental learning, particularly in classifying AI, human-generated, and various generative methods
    \item The use of pseudo-labeling to familiarize the model with augmentations in the test set, to enrich the training data, and to enable long-term learning that allows adaptation to small changes in generative models in real-world scenarios
\end{itemize}

In recent years, advancements in generative models have significantly enhanced the capabilities of AI in creating text and images. Large language models (LLMs) such as GPT-3 \cite{brown2020language}, GPT-4 \cite{achiam2023gpt}, and Mistral \cite{jiang2023mistral} have revolutionized text generation, enabling coherent and contextually rich outputs across diverse applications, but also raising challenges in detecting AI-generated text. Similarly, generative models like Stable Diffusion \cite{rombach2022high}, DALL-E \cite{ramesh2021zero}, Midjourney \cite{midjourney} have achieved remarkable success in producing high-quality, photorealistic images. However, the increasing realism of such outputs poses risks for misinformation and underscores the need for robust detection systems. Recent studies have explored ensemble methods combining multiple models \cite{mohamed2024proposed, abburi2023generative} or treating fake content as anomalies \cite{khalidOCFakeDectClassifyingDeepfakes2020}, achieving notable success in feature separability. However, these approaches often fall short in scalability when faced with the addition of new labels. Techniques like ArcFace have addressed some of these limitations, offering improved adaptability and performance in scenarios with evolving generative models. The remaining details and related works are provided in Appendix~\ref{sec:appendix_related_work}.

This paper is organized to provide a comprehensive overview of the proposed method and its evaluation. Section~\ref{sec:method} introduces the method, detailing its key components, including perceptual hashing, similarity measurement, and pseudo-labeling, while Section~\ref{sec:exp} presents the experimental setup, datasets, implementation details, and results to highlight the method's effectiveness. The paper concludes in Section~\ref{sec:conclude} with insights and future directions, with additional materials, including detailed related work, dataset descriptions, and supplementary experiments, provided in the Appendix (Sec.~\ref{sec:appendix}).

\section{Method}
\label{sec:method}
The proposed method comprises three main components: perceptual hashing (Sec. \ref{sec:hashing}), similarity measurement and comparison (Sec. \ref{sec:similarity}), and pseudo-labeling (Sec. \ref{sec:pseudo}). Each component is designed to address specific challenges in distinguishing between human-generated and AI-generated content, as well as classifying content from different generative models. 

\begin{figure*}[h]
    \centering
    \includegraphics[width=0.85\textwidth]{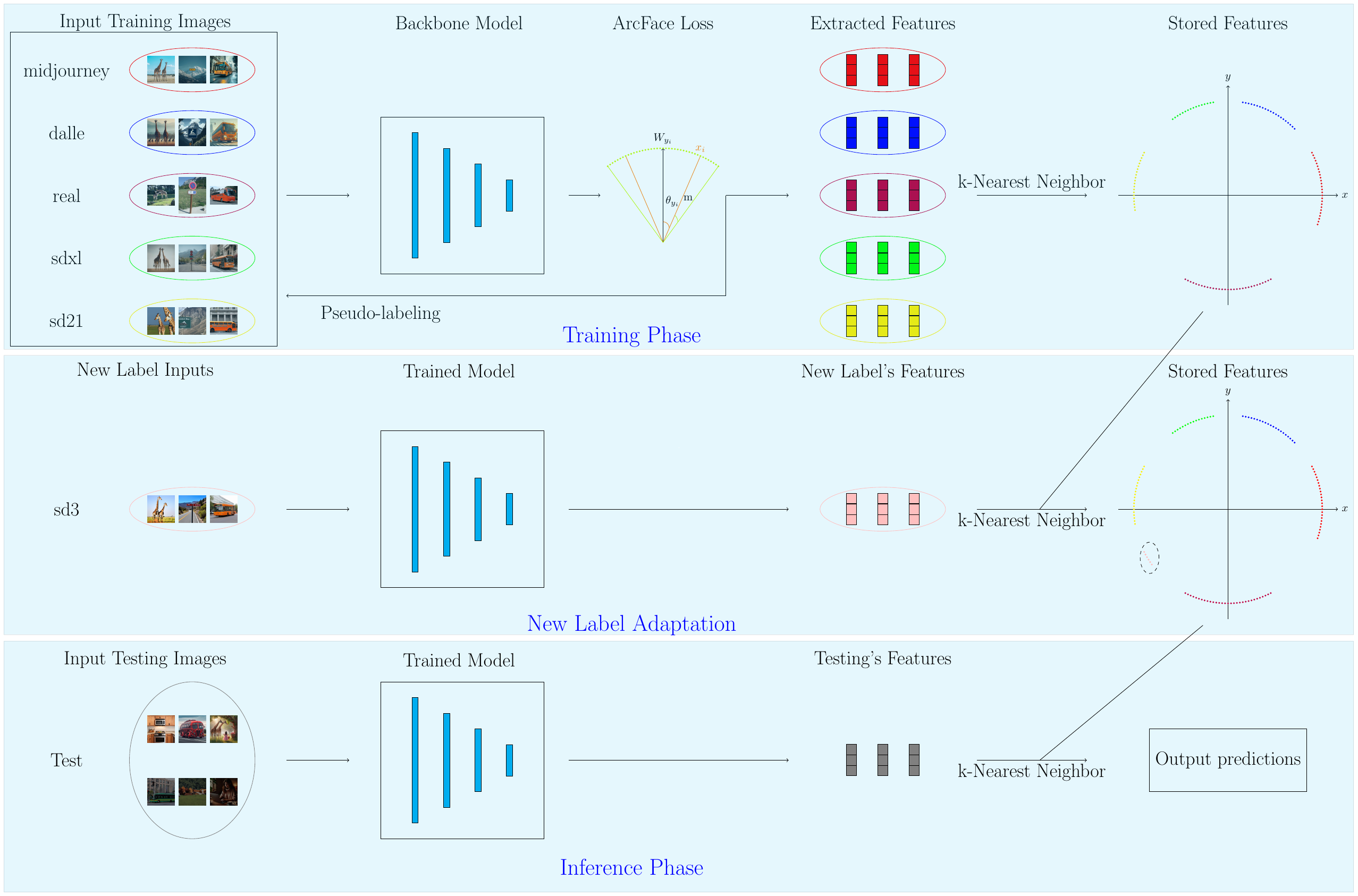}
    \caption{\small Overall framework of the proposed method, which consists of three stages, \ie, training, new label adaptation, and inference.}
    \label{fig:flow}
\end{figure*}

In Figure~\ref{fig:flow} we illustrate the overall framework, which consists of three stages: training, new label adaptation, and inference. In the training stage, we train the model using the ArcFace loss (Eq. \eqref{eq:arcface}) and apply pseudo-labeling to augment the dataset. Features are extracted from the input data and stored for later use, with pruning techniques for $k$-nearest neighbors employed to minimize storage requirements \cite{pedregosa2011scikit}. In the new label adaptation stage, features are extracted from the new data and seamlessly integrated into the existing feature storage. Finally, in the inference stage, features are extracted from incoming data and compared with the stored features to determine the most similar label. For text we adopted BART Large \cite{lewis2019bart}, and for image we used Swin Transformer V2 Base \cite{liu2022swin} as backbone model. This design ensures scalability and adaptability to new generative models while maintaining robust performance. 

\subsection{Perceptual hashing}
\label{sec:hashing}
Perceptual hashing is key to our method, offering a compact and efficient data representation for comparison. We train a model with ArcFace loss \cite{deng2019arcface} to enhance feature discrimination by enforcing class margins. The trained model outputs high-dimensional feature vectors, which serve as perceptual hash digests for the input samples.

Once trained, the model extracts features from all samples, converting them into high-dimensional feature vectors, referred to as hashing digests. These vectors are stored in a database and used for similarity comparisons. This approach enables efficient comparison and retrieval, leveraging the discriminative capability of the learned representations to distinguish between human-generated and AI-generated content, as well as between different generation models.

\subsection{Similarity measurement and comparison}
\label{sec:similarity}
To compare features, we employ a $k$-nearest neighbor ($k$-NN) approach, inspired by solutions from prior competitions \cite{toofanee2023dfu, jeon20201st}. This non-parametric method allows us to measure the similarity between samples effectively, leveraging the local neighborhood structure within the feature space. Specifically, we calculate the similarity metric using the cosine similarity, which is well-suited for high-dimensional feature spaces and ensures scale-invariant comparisons. The simplicity and adaptability of $k$-NN make it particularly suited for our scenario, where the inclusion of new labels or generative models is anticipated.

When new labels appear, our method avoids retraining the entire model. Instead, features from a few representative samples generated by the new AI model extracted and then appended to the existing feature set. Seamlessly integrating the new class into the similarity-based comparison process. This approach ensures a scalable and efficient adaptation mechanism, maintaining the system's flexibility to accommodate evolving AI-generated content without significant computational overhead.

\subsection{Pseudo labeling}
\label{sec:pseudo}
Pseudo-labeling is employed as a crucial component in our approach to address the limitations of a small labeled training dataset while improving the model's adaptability and robustness. By leveraging unlabeled data, pseudo-labeling effectively expands the training set, where high-confidence predictions of the model are used as surrogate labels \cite{cascante2021curriculum}. This augmentation strategy not only increases the diversity of the training samples but also familiarizes the model with potential augmentations present in test data, thereby enhancing generalization performance.

In deployment scenarios, pseudo-labeling demonstrates its practicality by exploiting the abundance of unlabeled data derived from user inputs. These inputs, typically available in substantial volumes and without explicit labels, provide a valuable resource for iterative model training and fine-tuning. Furthermore, as AI-generated content evolves with updates to generative models over time, pseudo-labeling allows the model to adapt dynamically to these changes. This adaptability ensures robust performance in real-world applications, where continuous learning from new and diverse data is essential to maintaining high accuracy and relevance.

Similar to the approach proposed in \cite{toofanee2023dfu}, we utilize a dynamic pseudo-labeling mechanism to enhance our training process. Specifically, at each training epoch, the model is used to predict labels for the test set, and only the top $p_{pseudo}$ percent of predictions with the highest confidence probabilities are selected as pseudo-labeled data. These high-confidence samples are then added to the training process to further refine the model.

In subsequent epochs, the pseudo-labeling step is repeated, and a new set of top $p_{pseudo}$ percent high-confidence predictions is selected. This iterative process helps to progressively eliminate potentially incorrect pseudo-labels, ensuring that only reliable samples contribute to the training. Importantly, the original labeled training data remains the primary focus during training, as it provides a reliable foundation with ground-truth labels. The pseudo-labeled samples constitute only a small fraction of the training data, serving as a complementary augmentation to improve model adaptability. The training continues in this manner until the model converges, leveraging the evolving predictions to adapt effectively and maintain robustness against noisy pseudo-labels.

\section{Experiments}
\label{sec:exp}
We conduct experiments to evaluate the effectiveness of our proposed method on both image and text datasets from the Defactify4 competition. The experiments are designed to test the model’s ability to distinguish between human-generated and AI-generated content, to classify content across different generative methods, and to adapt to new labels. Detailed descriptions of the datasets (Sec. \ref{sec:dataset}), implementation (Sec. \ref{sec:implementation}), and results (Sec. \ref{sec:results}) are provided in the following subsections.

\subsection{Dataset}
Our experiments utilize two benchmark datasets: \textbf{Defactify4-Image} and \textbf{Defactify4-Text}, each designed to evaluate AI vs. human classification and method-specific categorization. Both datasets comprise training and testing splits, with the testing sets further divided into unaltered data (\texttt{Test 1}) and augmented data (\texttt{Test 2}) to assess model robustness and generalization. \textbf{Defactify4-Image} consists of six classes, where one class represents real images derived from the COCO dataset \cite{lin2014microsoft}, and the remaining five classes correspond to outputs from different AI image-generation models. Meanwhile, \textbf{Defactify4-Text} comprises seven classes, including one class of human-written text and six classes generated by various text-generation models. Detailed descriptions of the datasets are provided in Appendix~\ref{sec:appendix_datasets}.

\subsection{Implementation}
\label{sec:implementation}
For the backbone model used to extract features from images, we adopt a Swin Transformer V2 Base model \cite{liu2022swin} with an image size of $256 \times 256$ and a window size of $8$, pre-trained on ImageNet \cite{deng2009imagenet}. For text data, we use a pre-trained BART Large model \cite{lewis2019bart} with a maximum token length of $512$. Training is done for $100$ epochs using Adam optimizer \cite{kingma2014adam} with a learning rate of $1 \times 10^{-4}$ and a batch size of $32$. In addition to the primary inputs, we incorporate auxiliary data that we found to be significant. For text data, we include the text length in terms of the number of characters and words as additional features. For image data, we leverage the image size as a supplementary input. After feature pooling from the backbone and concat with auxiliary data, we apply a fully connected layer with $512$ nodes to reduce the dimensionality of the features, using the Parametric ReLU (PReLU) function as activation layer.

For $k$-nearest neighbors ($k$-NN), we use the implementation provided by scikit-learn \cite{pedregosa2011scikit} with default parameters and adopt cosine as the metric, enabling faster computations and pruning to reduce the number of stored features. 
In the pseudo-labeling process, we select the top $p_{pseudo} = 5\%$ of predictions with the highest confidence probabilities for each predicted label $\hat{y}$, and assign $\hat{y}$ as the label for the corresponding data points.

Data augmentation is applied to both image and text data to enhance robustness. For text, we introduce random starting and ending positions in sequences and inject random meaningless strings at arbitrary points within the data. For image data, we employ augmentations including horizontal flip, Gaussian noise, image compression, and random brightness/contrast adjustments. To ensure compatibility with the input pipeline, resizing operations are performed while preserving the aspect ratio of the original image.

\subsection{Results}
\label{sec:results}
In this section, we present the results obtained from evaluating our proposed method on both image and text datasets. The competition for each data type is divided into two tasks: \texttt{Task A} focuses on distinguishing between AI-generated and human-produced content, while \texttt{Task B} classifies the content across different methods, with human-produced as one of the categories.

\begin{table}[h]
\begin{minipage}{.49\linewidth}
    \caption{\small Leaderboard of the \textbf{Defactify4-Image} task}
    \resizebox{0.7\linewidth}{!}{\begin{tabular}{l|ll}
    Team Name           & Task A           & Task B \\ \hline
    SeeTrails           & 0.8334           & 0.4986           \\
    NYCU                & 0.8329           & 0.491            \\
    random.py           & 0.8326           & 0.4936           \\
    Xiaoyu              & 0.8316           & 0.4888           \\
    TAHAKOM             & 0.8305           & 0.4816           \\
    SKDU                & 0.83             & 0.4864           \\
    Nitiz               & 0.8152           & 0.4193           \\
    OAR                 & 0.7996           & 0.2726           \\
    RoVIT               & 0.759            & 0.4222           \\ \hline
    dakiet (our method) & 0.833            & 0.4935          
    \end{tabular}}
    \label{tab:lb_image}
\end{minipage}
\begin{minipage}{.49\linewidth}
    \caption{\small Leaderboard of the \textbf{Defactify4-Text} task}
    \resizebox{0.7\linewidth}{!}{\begin{tabular}{l|ll}
    Team Name           & Task A           & Task B \\ \hline
    Sarang              & 1                & 0.9531           \\
    tesla               & 0.9962           & 0.9218           \\
    SKDU                & 0.9945           & 0.7615           \\
    Drocks              & 0.9941           & 0.627            \\
    Llama\_Mamba        & 0.988            & 0.4551           \\
    AI\_Blues           & 0.9547           & 0.3697           \\
    NLP\_great          & 0.9157           & 0.1874           \\
    Osint               & 0.8982           & 0.3072           \\
    Xiaoyu              & 0.803            & 0.5696           \\
    Rohan               & 0.7546           & 0.4053           \\ \hline
    dakiet (our method) & 0.9999           & 0.9082          
    \end{tabular}}
\label{tab:lb_text}

\end{minipage} 
\end{table}

In Table \ref{tab:lb_text} and Table \ref{tab:lb_image}, we present the leaderboards (evaluated on \texttt{Test 2}) for the text and image tasks, respectively, in the competition. Our proposed method demonstrates competitive performance compared to other teams. Specifically, we achieved the 2nd place in Task A and the 3rd place in Task B for both text and image datasets. This highlights the effectiveness of our method, which not only achieves high accuracy but also incorporates the ability for continual learning through pseudo-labeling. More importantly, our method has the potential to expand the number of labels, accommodating new classes in the future. This is crucial given the increasing diversity of AI-generated models and outputs. The capability to seamlessly add new labels ensures our approach remains adaptable and robust as the landscape of generative AI continues to evolve.

\section{Conclusion}
\label{sec:conclude}
This work presents a robust and scalable framework for the detection and classification of AI-generated content across text and image domains. By integrating perceptual hashing, similarity measurement, and pseudo-labeling, our method addresses the critical challenges posed by the rapid evolution of generative AI. Experimental results on the \texttt{Defactify4} dataset validate the effectiveness of our approach, which achieves competitive performance while maintaining adaptability to new generative models. Importantly, the proposed method provides a flexible solution for continual learning, accommodating the growing diversity of AI-generated content without the need for retraining. This adaptability ensures relevance in real-world applications, where the landscape of generative AI is dynamic and ever-expanding. Future work includes exploring techniques to convert high-dimensional floating-point vectors into compact binary representations while maintaining classification performance.

\bibliography{references}

\section{Appenddix}
\label{sec:appendix}
\subsection{Related work}
\label{sec:appendix_related_work}
In this section, we review related work on three key areas. First, we discuss 
loss functions used for classification tasks (Sec. \ref{sec:lmcl}). 
Next, we explore advancements in text generation with large language models (LLMs), which have significantly advanced the field of natural language processing, and the detection of AI-generated text (Sec. \ref{sec:gen_text}). Finally, we present recent progress in image generation models and the detection of AI-generated images (Sec. \ref{sec:gen_image}).

\subsubsection{Loss functions for classification}
\label{sec:lmcl}
We begin by analyzing the conventional Softmax loss, a widely used loss function in classification tasks:

\begin{equation}
\label{eq:softmax}
    L_{\text{Softmax}}=-\frac{1}{N}\sum\limits_{i=1}^{N}{\log }\frac{{{e}^{W_{{{y}_{i}}}^{T}{{x}_{i}}+{{b}_{{{y}_{i}}}}}}}{\sum\limits_{j=1}^{n}{{{e}^{W_{j}^{T}{{x}_{i}}+{{b}_{j}}}}}}
    =
    -\frac{1}{N}\sum\limits_{i=1}^{N}{\log }\frac{{{e}^{\left\| {W_{{{y}_{i}}}} \right\|\left\| {{x}_{i}} \right\|\cos \left( {{\theta }_{y_i x_i}} \right)+{{b}_{{{y}_{i}}}}}}}{\sum\limits_{j=1}^{n}{{{e}^{\left\| {W_j} \right\|\left\| {{x}_{i}} \right\|\cos \left( {{\theta }_{j x_i}} \right)+{{b}_{j}}}}}}
\end{equation}
where ${{y}_{i}}$ is the class label of the $i^{\text{th}}$ sample, ${{\theta }_{j x_i}}$ is the angle between the weight ${{W}_{j}}^{T}$ and the feature ${{x}_{i}}$, $n$ is the total number of classes, and $N$ is the number of samples. Models trained with the Softmax loss are limited to a fixed set of classes and require retraining whenever new labels are added. This limitation comes from the fully connected layer, which has a fixed number of nodes corresponding to the predefined classes.

One workaround for models trained with Softmax loss is to remove the final fully connected layer and use the extracted feature embeddings. However, as shown in \cite{deng2019arcface}, this method still has limitations. While Softmax can generate separable feature embeddings, it often leads to unclear decision boundaries, making it ineffective for robust classification without further adjustments. In contrast, the Large Margin Cosine Loss (LMCL) explicitly introduces a margin between the closest classes, leading to better-defined decision boundaries and more robust separability.

By applying $l_{2}$ normalization to both the weights and features, ensuring $\left\| {{W}_{j}} \right\|=\left\| {{x}_{i}} \right\|=1$, introducing a scaling factor $s$, and setting the bias ${{b}_{j}}=0$, as described in \cite{wang2017normface}, the original Softmax loss function (\ref{eq:softmax}) is reformulated as:

\begin{equation}
\label{eq:normface}
    L_{NormFace}=-\frac{1}{N}\sum\limits_{i=1}^{N}{\log }\frac{{{e}^{s\cos {{\theta }_{{{y}_{i}}}}}}}{\sum\limits_{j=1}^{n}{{{e}^{s\cos {{\theta }_{ji}}}}}}=-\frac{1}{N}\sum\limits_{i=1}^{N}{\log }\frac{{{e}^{s\cos {{\theta }_{{{y}_{i}}}}}}}{{{e}^{s\cos {{\theta }_{{{y}_{i}}}}}}+\sum\limits_{j=1,j\ne {{y}_{i}}}^{n}{{{e}^{s\cos {{\theta }_{ji}}}}}}.
\end{equation}

As mentioned earlier, the Softmax loss function suffers from limitations when it comes to handling the addition of new classes and ensuring clear decision boundaries. Since the embedding features are distributed around each class center on the hypersphere, an additive angular margin penalty, $m$, between the feature vector $x_i$ and the corresponding class weight $W_{y_i}$ can be introduced. This margin enhances both the intra-class compactness and inter-class separability, encouraging better feature discrimination. In other words, the model learns to increase the angular distance between different classes while simultaneously reducing the distance within the same class.

\begin{equation}
\label{eq:arcface}
    L_{ArcFace}=-\frac{1}{N}\sum\limits_{i=1}^{N}{\log }\frac{{{e}^{s\left( \cos \left( {{\theta }_{{{y}_{i}}}}+m \right) \right)}}}{{{e}^{s\left( \cos \left( {{\theta }_{{{y}_{i}}}}+m \right) \right)}}+\sum\limits_{j=1,j\ne {{y}_{i}}}^{n}{{{e}^{s\cos {{\theta }_{ji}}}}}}
\end{equation}

This reformulation, known as ArcFace \cite{deng2019arcface}, incorporates the additive angular margin $m$ directly into the decision boundaries, as shown in Equation (\ref{eq:arcface}). ArcFace has established itself as a state-of-the-art approach for classification tasks, especially in scenarios where new labels may emerge outside the training set. While newer loss functions may surpass ArcFace in specific scenarios, its widespread use in competitions and ease of implementation have solidified its position as a benchmark choice for its combination of performance, versatility, and reliability \cite{jeon20201st}.

\subsubsection{Detection of AI-generated text}
\label{sec:gen_text}
Text generation has witnessed rapid advancements with the development of large language models (LLMs) such as GPT-3 \cite{brown2020language}, GPT-4 \cite{achiam2023gpt}, and Mistral \cite{jiang2023mistral}. These models are designed to generate coherent and contextually relevant text, leveraging vast pretraining datasets to produce outputs that closely resemble human-written content. Applications of these models include creative writing, summarization, and conversational agents, showcasing their ability to adapt across diverse linguistic tasks. 

One notable advantage of LLMs lies in their capacity to generalize across tasks with minimal fine-tuning, often performing well with zero-shot or few-shot learning. This flexibility has enabled their widespread adoption in various fields, but it has also introduced challenges in distinguishing between human-written and AI-generated text. As models become increasingly realistic, research has focused on methods to identify synthetic text, emphasizing the importance of robust detection frameworks to mitigate misuse and ensure trust in AI-generated content \cite{wu2024continual}. Recent studies have explored ensemble methods that aggregate output probabilities from multiple backbone models trained with softmax loss, achieving competitive results in text classification tasks \cite{mohamed2024proposed, abburi2023generative}. However, while effective, these approaches exhibit limited scalability when faced with a growing number of labels, underscoring the need for alternative methods capable of adapting to the dynamic nature of generative AI.

\subsubsection{Detection of AI-generated images}
\label{sec:gen_image}
Recent years have seen remarkable progress in image generation, driven by models such as Stable Diffusion \cite{rombach2022high}, DALL-E \cite{ramesh2021zero}, Midjourney \cite{midjourney}.
These generative models are capable of creating high-quality, photorealistic images from textual prompts, offering unprecedented control over the content and style of the generated visuals. Their applications range from digital art and design to content creation and entertainment, making them invaluable tools in creative industries.

These models operate by learning to generate images that align with textual descriptions, often leveraging latent space representations to ensure fine-grained control over image attributes. The rapid evolution of these models has enabled the creation of visually compelling and contextually accurate outputs, but it has also raised ethical and practical concerns. The ability to generate highly realistic images poses risks for misinformation and deceptive media. Consequently, the development of reliable classification systems to differentiate between real and AI-generated images has become a pressing need to ensure the ethical use of such technologies \cite{deandres2024frcsyn}.

Deep learning generated fake images are typically created by generative neural network (GAN) models, but can also be created using autoencoders \citep{wangCNNGeneratedImagesAre2020}. In most cases, the creation of fake images consists of replacing a person or a face in an existing image or video with another person or face.
As videos contain more information than images, most methods apply to the detection of fake videos \citep{camachoComprehensiveReviewDeepLearningBased2021}. 
However some methods have been proposed for images.
The first methods for fake image detection focus on the detection of images generated by a specific GAN \citep{marraDetectionGANGeneratedFake2018}. 
Common CNN models detect spatial cues such as artifacts on facial boundaries \citep{liFaceXRayMore2020}, or traces left by the GAN  \citep{liExposingDeepFakeVideos2018,xuanGeneralizationGANImage2019,wangCNNGeneratedImagesAre2020}.
\citet{khalidOCFakeDectClassifyingDeepfakes2020} propose a one-class variational autoencoder model to detect fake images as anomalies.
However, these methods cannot adapt to new AI image generation models.

\subsection{Additional experiments}
In this section, we present additional experiments to evaluate the adaptability of our method when integrating a new label without retraining the model. 
\begin{table}[h]
\caption{Performance results on the \textbf{Defactify4-Image} dataset for Task 1 (AI-generated vs. Human-produced) and Task 2 (Classification of different methods). \texttt{all} indicates using the entire dataset with the proposed method (using ArcFace loss), while \texttt{all-x} refers to the dataset excluding samples from the class labeled as \texttt{x}. And \texttt{swinv2's training data} refers to using a pre-trained Swin Transformer V2 \cite{liu2022swin} model to extract features from images without additional training.}
\resizebox{\linewidth}{!}{
\begin{tabular}{cc|cccccccccc}
\multicolumn{2}{c|}{\multirow{3}{*}{Testing data}} & \multicolumn{10}{c}{Training data}                                                                                                                                                                                                                         \\ \cline{3-12} 
\multicolumn{2}{c|}{}                              & \multicolumn{1}{c|}{\multirow{2}{*}{\begin{tabular}[c]{@{}c@{}}swinv2's\\ training data\end{tabular}}} & \multicolumn{9}{c}{Defactify dataset}                                                                                                               \\ \cline{4-12} 
\multicolumn{2}{c|}{}                              & \multicolumn{1}{c|}{}                                                                                & \multicolumn{1}{c|}{all-dalle-sdxl} & all-midjourney & all-dalle & all-real & all-sdxl & all-sd21 & \multicolumn{1}{c|}{all-sd3} & Softmax & all    \\ \hline
\multirow{2}{*}{Test 1}          & Task A          & \multicolumn{1}{c|}{0.98}                                                                            & \multicolumn{1}{c|}{0.98}           & 1.00           & 1.00      & 0.98     & 0.99     & 1.00     & \multicolumn{1}{c|}{0.99}    & 1.00    & 1.00   \\
                                 & Task B          & \multicolumn{1}{c|}{0.90}                                                                            & \multicolumn{1}{c|}{0.92}           & 1.00           & 0.99      & 0.94     & 0.98     & 0.99     & \multicolumn{1}{c|}{0.96}    & 1.00    & 1.00   \\ \hline
\multirow{2}{*}{Test 2}          & Task A          & \multicolumn{1}{c|}{0.7651}                                                                          & \multicolumn{1}{c|}{0.7817}         & 0.8295         & 0.8306    & 0.8034   & 0.8278   & 0.8293   & \multicolumn{1}{c|}{0.8250}  & 0.8303  & 0.8330 \\
                                 & Task B          & \multicolumn{1}{c|}{0.3170}                                                                          & \multicolumn{1}{c|}{0.3598}         & 0.4506         & 0.4766    & 0.3920   & 0.4672   & 0.4637   & \multicolumn{1}{c|}{0.4383}  & 0.4870  & 0.4935 \\ \hline
\end{tabular}}
\label{tab:result_image}
\end{table}

Table \ref{tab:result_image} shows the performance of our method on the \textbf{Defactify4-Image} dataset. In both tests, both the Softmax loss and our proposed framework (using ArcFace loss) perform well and have relatively similar results. Our proposed method performs slightly better in \texttt{Test 2}, and both methods achieve perfect accuracy in \texttt{Test 1}. Using the pre-trained model without fine-tuning results in relatively low performance. This is because the model was trained on a broad dataset and does not perform well on a fine-grained dataset for specific tasks. Additionally, AI-generated images may have a different distribution compared to the images the model was trained on, causing difficulties in performance. When removing two labels from the data, the model experiences a significant drop in performance, though it still outperforms the pre-trained Swin Transformer V2 Base model. When removing one label for training, the model still maintains acceptable performance, except when excluding the "real" label. The reason for this is that the gap between AI-generated images and human-produced images is relatively large, so training exclusively on AI-generated images causes the model to become unfamiliar with real image data. However, when removing one AI-generated label, the model performs slightly lower but still remains highly competitive compared to the Softmax method. This demonstrates the potential for future expansion of the number of labels in our proposed method.

\begin{table}[h]
\caption{Performance results on the \textbf{Defactify4-Text} dataset for Task 1 (AI-generated vs. Human-produced) and Task 2 (Classification of different methods). \texttt{all} indicates using the entire dataset with the proposed method (using ArcFace loss), while \texttt{all-x} refers to the dataset excluding samples from the class labeled as \texttt{x}. And \texttt{BART's training data} refers to using a pre-trained BART \cite{lewis2019bart} model to extract features from input text without additional training.}
\resizebox{\linewidth}{!}{
\begin{tabular}{cc|ccccccccccc}
\multicolumn{2}{c|}{\multirow{3}{*}{Testing data}} & \multicolumn{11}{c}{Training data}                                                                                                                                                                                                                                                     \\ \cline{3-13} 
\multicolumn{2}{c|}{}                              & \multicolumn{1}{c|}{\multirow{2}{*}{\begin{tabular}[c]{@{}c@{}}BART's\\ training data\end{tabular}}} & \multicolumn{10}{c}{Defactify dataset}                                                                                                                                            \\ \cline{4-13} 
\multicolumn{2}{c|}{}                              & \multicolumn{1}{c|}{}                                                                              & \multicolumn{1}{c|}{all-yi-llama} & \multicolumn{1}{l}{all-Human} & all-gemma2 & all-mistral & all-qwen2 & all-llama & all-yi & \multicolumn{1}{c|}{all-gpt4o} & Softmax & all    \\ \hline
\multirow{2}{*}{Test 1}          & Task A          & \multicolumn{1}{c|}{0.98}                                                                          & \multicolumn{1}{c|}{0.99}         & 0.98                          & 1.00       & 1.00        & 0.99      & 1.00      & 1.00   & \multicolumn{1}{c|}{1.00}      & 1.00    & 1.00   \\
                                 & Task B          & \multicolumn{1}{c|}{0.94}                                                                          & \multicolumn{1}{c|}{0.95}         & 0.95                          & 0.96       & 0.96        & 0.95      & 0.95      & 0.96   & \multicolumn{1}{c|}{0.95}      & 0.96    & 0.96   \\ \hline
\multirow{2}{*}{Test 2}          & Task A          & \multicolumn{1}{c|}{0.9970}                                                                        & \multicolumn{1}{c|}{0.9978}       & 0.9950                        & 0.9980     & 0.9980      & 0.9970    & 0.9978    & 0.9971 & \multicolumn{1}{c|}{0.9947}    & 0.9963  & 0.9999 \\
                                 & Task B          & \multicolumn{1}{c|}{0.7946}                                                                        & \multicolumn{1}{c|}{0.8190}       & 0.8350                        & 0.8738     & 0.8692      & 0.8206    & 0.8462    & 0.8433 & \multicolumn{1}{c|}{0.8268}    & 0.9049  & 0.9082 \\ \hline
\end{tabular}}
\label{tab:result_text}
\end{table}

Table \ref{tab:result_text} shows the performance of our method on the \textbf{Defactify4-Text} dataset. In both tests, similar to the image task the Softmax method and our proposed method perform well and have relatively similar results. Our proposed method performs slightly better in \texttt{Test 2}. Using the pre-trained model without fine-tuning results in relatively low performance. When removing one label for training, the model still maintains acceptable performance. However, compared to the image dataset, the performance drops when using only the pre-trained BART Large model or excluding the "human" label is less noticeable here. This suggests that AI-generated text has a distribution that is not drastically different from the original wide dataset, and human-generated text does not deviate much from the other labels. This indicates that text generation models are performing better than image generation models, as the distribution of outputs between human and AI-generated text is less distinct. However, more data is needed to fully validate this observation. When removing two labels from the data, the model experiences a significant drop in performance, though it still outperforms the pre-trained model.

In both Table \ref{tab:result_image} and Table \ref{tab:result_text}, when removing one AI-generated label the performance of the model decreases in both tasks. However, the model performs slightly lower but still remains highly competitive compared to the Softmax method. This demonstrates the potential for future expansion of the number of labels in our proposed method.

\subsection{Dataset detail}
\label{sec:appendix_datasets}
The experiments are conducted on two benchmark datasets: \textbf{Defactify4-Image} (Sec. \ref{sec:data_img}) and \textbf{Defactify4-Text} (Sec. \ref{sec:data_text}). Each dataset includes training and testing splits tailored for two tasks: AI vs. human classification and method-specific categorization. The datasets are carefully designed with a mix of real and AI-generated data, incorporating augmentations to test the model’s robustness and generalization capabilities. Below, we provide detailed descriptions of each dataset.
\label{sec:dataset}
\subsubsection{Dataset of AI-generated images}
\label{sec:data_img}
The \textbf{Defactify4-Image} dataset is a benchmark designed to evaluate the ability to distinguish between real and AI-generated images. It comprises seven data categories (6 classes and captions),  including captions and various image sources. Among these, real images selected from the COCO dataset \cite{lin2014microsoft} are represented by the \texttt{coco\_image} class. While the other five categories (\texttt{sd3\_image}, \texttt{sd21\_image}, \texttt{sdxl\_image}, \texttt{dalle\_image}, \texttt{midjourney\_image}) are generated using specific AI models: Stable Diffusion (v3 \cite{esser2024scaling}, v2.1 \cite{rombach2022high}, XL \cite{podell2023sdxl}), DALL-E \cite{betker2023improving}, and MidJourney \cite{midjourney}, respectively. Captions act as the input prompts for these generative models and correspond to the caption of real images in the dataset.

\paragraph{Training Data}
The training set consists of $42,000$ images across six classes, with $7,000$ samples per class. Each class corresponds to one of the five generative models and the real image class (\texttt{coco\_image}). All images share the same caption within their index, for instance, \texttt{sd3\_image[i]}, \texttt{sd21\_image[i]} are generated from the same \texttt{caption[i]} of \texttt{coco\_image[i]}. Where \texttt{record[i]} refers to the $i^{th}$ sample of the $record$ in the dataset.

\paragraph{Testing Data} The dataset provides two distinct test sets to assess model performance. \texttt{Test 1} comprises $9,000$ images, each paired with its original caption, representing unaltered outputs from generative models or real images. While \texttt{Test 2} consists of $45,000$ images where augmentation techniques have been applied, enabling evaluation of the model's robustness and generalization across various transformations.

\subsubsection{Dataset of AI-generated text}
\label{sec:data_text}
The \textbf{Defactify4-Text} dataset is a benchmark designed to evaluate the ability to distinguish between human-written and AI-generated text. It comprises eight columns: \texttt{prompt}, \texttt{Human\_story}, \texttt{gemma-2-9b}, \texttt{mistral-7B}, \texttt{qwen-2-72B}, \texttt{llama-8B}, \texttt{yi-large}, and \texttt{GPT\_4-o}. The \texttt{prompt} column contains the instruction, while \texttt{Human\_story} is a human-written text corresponding to the \texttt{prompt}. The remaining columns represent outputs from various generative models, including \texttt{gemma-2-9b} \cite{gemma_2024}, \texttt{mistral-7B} \cite{jiang2023mistral}, \texttt{qwen-2-72B} \cite{qwen2}, \texttt{llama-8B} \cite{llama3modelcard}, \texttt{yi-large} \cite{young2024yi}, and \texttt{GPT\_4-o} \cite{achiam2023gpt}, based on the provided \texttt{prompt}.

\paragraph{Training Data}
The training set consists of $51,248$ text samples across seven classes, with $7,321$ samples per class. Each class corresponds to one of the generative models or the human-written text class (\texttt{Human\_story}). Similar to \textbf{Defactify4-Image} all samples share the same prompt within their index.

\paragraph{Testing Data}
For evaluation, the dataset provides two separate test sets. \texttt{Test 1} includes $10,983$ samples, each associated with its original prompt and the corresponding generated or human-written text. These samples represent the raw outputs from the generative models or human authors. On the other hand, \texttt{Test 2} contains $10,963$ samples where various augmentation techniques have been applied to the text data, allowing for the assessment of model robustness and generalization under various transformations.

\subsection{Discussion}
The proposed method demonstrates strong scalability, particularly in its ability to adapt to new labels and modalities. However, the approach has certain limitations, primarily the need to store feature representations. Specifically, the extracted features consist of high-dimensional floating-point vectors, which can impose significant memory requirements as the dataset size increases. For instance, in our implementation, each sample in the training set requires $32 \times 512 = 16,384$ bits to store, given that each floating-point number occupies $32$ bits, and $512$ is the length of the vector. This storage demand becomes challenging with large-scale datasets.

Recent studies have highlighted strategies to address this issue by converting high-dimensional floating-point vectors into shorter binary vectors \cite{cui2020exchnet, shen2022semicon}. These approaches significantly reduce storage requirements but often come with a trade-off in terms of performance, making them less suitable for competitive tasks such as those in this challenge. Exploring methods to balance memory efficiency and accuracy remains an important area for future research.

\end{document}